\documentclass{article}
\usepackage{spconf,amsmath,graphicx}
\usepackage{multirow}
\usepackage{bbding}
\usepackage{stfloats}
\usepackage{pifont}
\usepackage{hyperref}
\usepackage{booktabs} 
\ninept 

\title{DIFFDUB: PERSON-GENERIC VISUAL DUBBING USING INPAINTING RENDERER WITH DIFFUSION AUTO-ENCODER}

\name{Tao Liu$^1$, Chenpeng Du$^1$, Shuai Fan$^2$, Feilong Chen$^2$, $^{\dagger}$Kai Yu$^1$\thanks{$^{\dagger}$Kai Yu is the corresponding author.}}
\address{$^1$MoE Key Lab of Artificial Intelligence, AI Institute \\ $^1$X-LANCE Lab, Shanghai Jiao Tong University
\\$^2$AISpeech Ltd, Suzhou China}
\begin{document}

\maketitle


\begin{abstract}



Generating high-quality and person-generic visual dubbing remains a challenge. Recent innovation has seen the advent of a two-stage paradigm, decoupling the rendering and lip synchronization process facilitated by intermediate representation as a conduit. Still, previous methodologies rely on rough landmarks or are confined to a single speaker, thus limiting their performance.
In this paper, we propose \textbf{\em{DiffDub}}: \textbf{Diff}usion-based \textbf{dub}bing. We first craft the Diffusion auto-encoder by an inpainting renderer incorporating a mask to delineate editable zones and unaltered regions. This allows for seamless filling of the lower-face region while preserving the remaining parts.
Throughout our experiments, we encountered several challenges. Primarily, the semantic encoder lacks robustness, constricting its ability to capture high-level features. Besides, the modeling ignored facial positioning, causing mouth or nose jitters across frames. To tackle these issues, we employ versatile strategies, including data augmentation and supplementary eye guidance. Moreover, we encapsulated a conformer-based reference encoder and motion generator fortified by a cross-attention mechanism. This enables our model to learn person-specific textures with varying references and reduces reliance on paired audio-visual data.
Our rigorous experiments comprehensively highlight that our ground-breaking approach outpaces existing methods with considerable margins and delivers seamless, intelligible videos in person-generic and multilingual scenarios.

\end{abstract}
\section{Introduction}
\label{sec:intro}

\begin{figure}[!pt]
    \centering
    \includegraphics[width=1.0\linewidth]{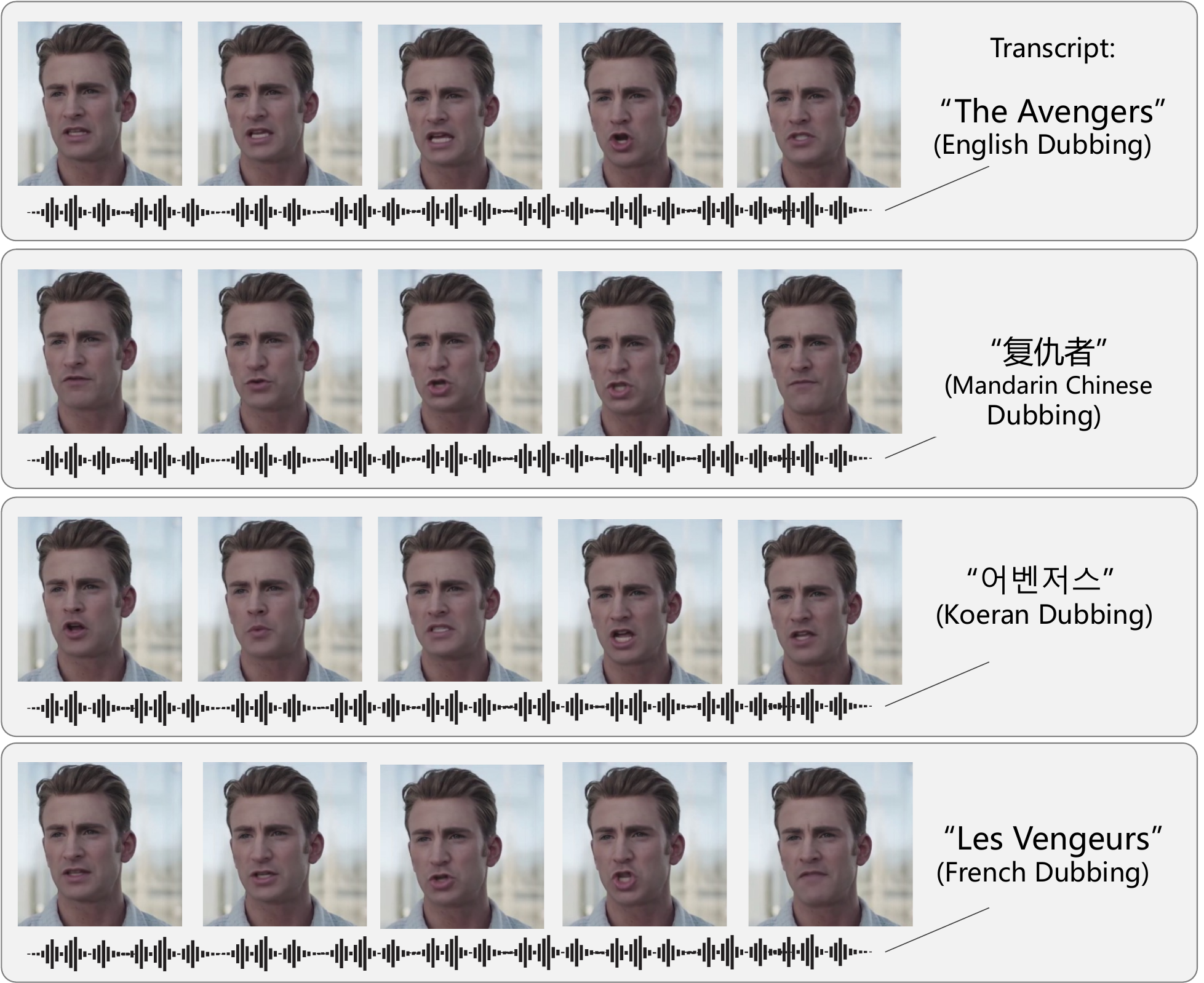}
    \caption{\textbf{Dubbed videos with audio in various languages}. Our method can produce seamless and intelligible videos. }
    \label{fig:demo}
\end{figure}

Visual dubbing~\cite{wav2lip, xie2021towards, dinet}, an area intrinsically linked to talking head synthesis, requires the meticulous alignment of lower-face movements in a source video with corresponding driving audio, while preserving the original identity, head pose, and background depicted in the source video. The utility of this task becomes evident when there is a need to modify or substitute the audio content with alternative audio, typically for translation purposes~\cite{kr2019towards}, as indicated in Figure~\ref{fig:demo}.

The task navigates through multiple challenges\cite{sheng2022deep}: maintaining high {\bf visual quality}, ensuring {\bf temporal consistency}, and perfecting {\bf lip synchronization}. Visual quality necessitates the seamless alteration of the lower facial area and a harmonious integration of the generated region with the unaltered part of the image. Temporal consistency preserves a fluid and natural motion between successive video frames, preempting jitters, or abrupt transitions. Lip synchronization mandates effective alignment between audio and visual cues. A harmonious balance between temporal consistency and lip synchronization contributes to the intelligibility of the video\cite{mcgurk1976hearing}.

Regarding visual quality, a large proportion of methods~\cite{wav2lip, zhou2020makelttalk, dinet} deploy Generative Adversarial Networks~(GANs) as the rendering network. Despite this, these approaches grapple with challenges such as training instability and mode collapse~\cite{dhariwal2021diffusion}.
The Diffusion Denoising Probabilistic Model (DDPM)~\cite{ho2020denoising} stands out as a probabilistic generative model that has notably displayed promising visual quality within the realm of talking head synthesis~\cite{difftalk, stypulkowski2022diffused, bigioi2023speech, daetalker}. The foundation of the diffusion model rests on the iterative addition (diffusion process) and removal (denoising process) of noise. Despite DDPM's considerable achievements in talking face generation~\cite{difftalk, stypulkowski2022diffused, bigioi2023speech}, progress has been limited within the sphere of visual dubbing, where only the lower facial area requires modification while the rest of the image remains unaltered. This stipulation concurs with image in-painting~\cite{lugmayr2022repaint, avrahami2022blended}, where alterations are confined to the repaired region. The modified segment must blend flawlessly with the original image to emanate a natural and coherent image. However, their primary focus on static images and an absence of consideration for temporal consistency. 




Concerning temporal consistency and lip synchronization, current methods \cite{wav2lip, pcavs, dinet} have yet to fully leverage the perks of sequence modeling or transduction methods, such as Transformer \cite{vaswani2017attention} or Conformer \cite{gulati2020conformer}. These approaches typically operate on short audio clips, e.g., 200 ms in Wav2Lip~\cite{wav2lip}. This duration, however, might fall short for some phonemes~\cite{igras2013length}, let alone their combinations. Relying on such brief audio clips introduces several challenges, including needing copious, aligned audio-visual data to achieve satisfactory results. Moreover, this approach regularly results in lip-synced but unintelligible videos~\cite{wang2023seeing} that often exhibit brisk lip movements or exaggerated mouth articulations~\cite{wav2lip, pcavs}. To surmount these challenges and harness the potential of sequence modeling, recent strategies~\cite{fan2022faceformer, chen2023improving, iplap, daetalker} have devised two-stage networks composed of an audio-to-representation generator and a representation-to-video rendering network. The generator, often reliant on a transformer model, maps the audio sequence onto a representative sequence while the rendering network fabricates the final video based on this representation. These methods, by decoupling the rendering process, successfully capture long-range dependencies. DAE-talker~\cite{daetalker} advances this approach by exploiting semantic latent variables from Diffusion Autoencoders~(Diff-AE) \cite{diffae} as opposed to employing explicit structural representations like blendshapes~\cite{fan2022faceformer, chen2023improving} or landmark coefficients~\cite{iplap}. Diff-AE can be viewed as a data compression or a representation learner by encoding an input image into a semantic latent variable and reconstructing the image from this latent space. Building upon this capability and capitalizing on Diff-AE's demonstrated competence in learning meaningful representations, DAE-talker achieves impressive outcomes. However, its applicability remains confined to a single speaker, like Obama, which dampens its generalization capacity.

\begin{figure*}[!pt]
    \centering
    \includegraphics[width=1.0\linewidth]{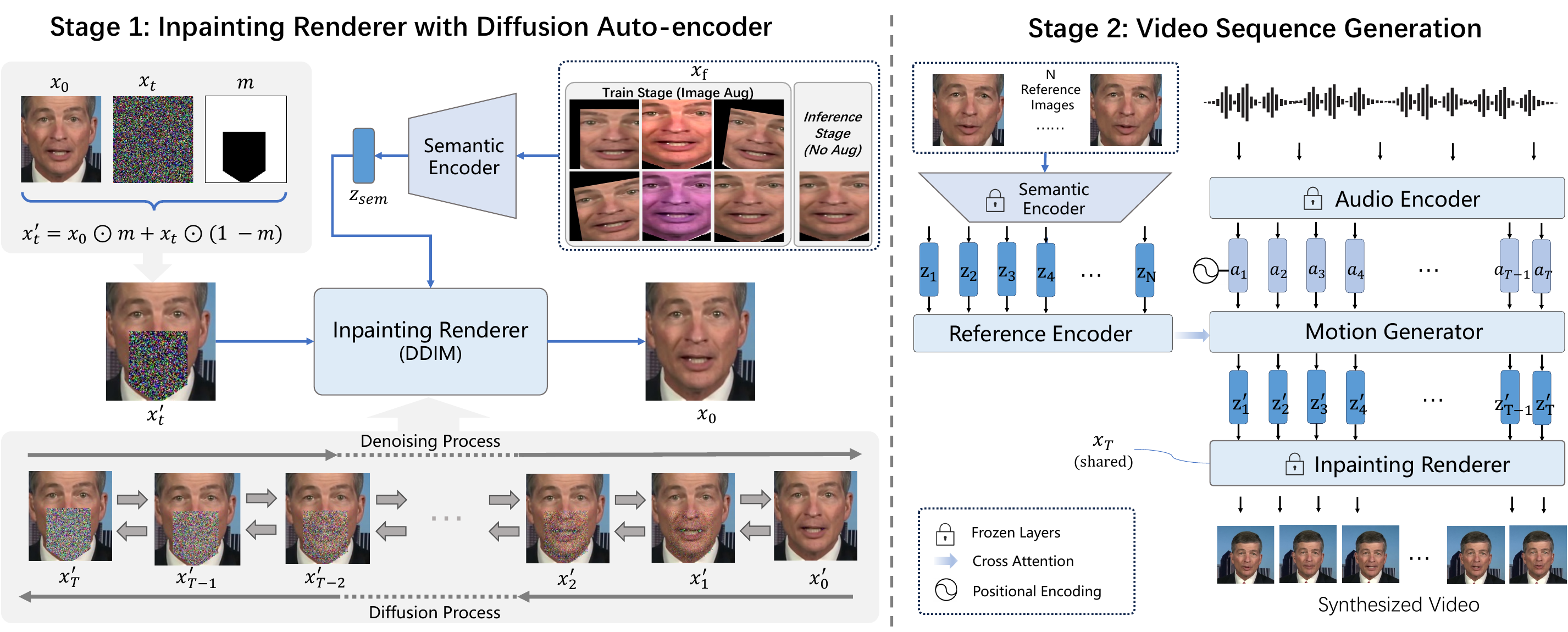}
    \caption{\textbf{Architecture of DiffDub}. Our DiffDub approach upholds a two-stage paradigm encompassing {\em inpainting rendering with Diffusion Auto-encoder} and {\em video sequence generation}. In the first stage, we usher in a Diffusion Auto-encoder with masked conditions to generate semantic latent codes $z$ through the semantic encoder. Subsequently, during the video generation phase, the semantic latent code $z$, in tandem with the audio latent code $a$ derived from an extant model, is employed to generate the final videos.}
    \label{fig:arch_graph}
\end{figure*}


In a bid to holistically address the aforementioned challenges, we introduce \textbf{\em{DiffDub}}: a person-generic visual dubbing methodology underpinned by DDPM. This paper encapsulates several contributions itemized below. \begin{itemize} \item We have designed a potent inpainting renderer tasked with the generation of the modifiable region under the supervision of immutable components and semantic conditioning. This method can generate seamlessly blended lower facial regions that far surpass the capabilities of preceding methods. \item We have proposed diverse strategies aimed at bolstering the robustness of the semantic encoder. These tactics enable the encoder to apprehend subtle movements and supply meticulous positional data for the mouth and nose regions. \item By leveraging the Conformer model with cross-attention, we have successively adapted our methodology to cater to an assortment of references and audio sequences. This adaptation allows our model to assimilate person-specific textures while reducing reliance on paired audio-visual data. \item We have performed thorough quantitative and qualitative evaluations in both few-shot and one-shot settings on URL\footnote{ \url{https://liutaocode.github.io/DiffDub/}}. \end{itemize}

\section{DiffDub Framework}
\label{sec:methods}

\subsection{Inpainting Renderer with Diffusion Auto-encoder}

This module is responsible for inpainting rendering and latent semantic representation learning. Before diving into its details, it is crucial to briefly recapitulate the noising and denoising process inherent to vanilla diffusion~\cite{ho2020denoising}.
Given a data distribution $x_0 \sim q\left(x_0\right)$, the noising process produces a series of latent $x_\text{1}, \ldots, x_\text{T}$ by adding Gaussian noise with variance at each time $t$. For the denoising process, we use $\mathcal{N} ( \mu_\theta ( x_t, t), \sigma_t )$ to model $q\left(x_\text{t-1}|x_\text{t}\right)$. Thus, we can train a deep neural network $p_\theta$ to predict the mean of Gaussian noise. However, the vanilla diffusion is not controllable, and we define two guidance information for the dubbing task by segregating the full image into two segments using a lower-face mask. The first segment is the reference $x_{0}$, providing the information to be retained, including identity, pose, and background. The second segment is the facial motion $x_{f}$, supplying the information for the editable region. The mask encompasses the lower facial area and can be procured through a landmark predictor \cite{bulat2017far}.

Our proposed approach is comprised of a semantic encoder $z_{\text{sem}} = \operatorname{Enc}_\phi\left(x_{\text{f}}\right)$ and an inpainting renderer $p\left(x^{\prime}_{t-1} \mid x^{\prime}_t, z_{\text{sem}}\right)$. Compared to Diff-AE~\cite{diffae}, the input of the semantic encoder is solely the facial area $x_f$ rather than a full image $x_0$. We opt for this approach to ensure that the semantic encoder solely imparts facial motion information. We also incorporate image augmentation throughout the training phase to ascertain the learning of high-level features by the semantic encoder instead of trivial patterns. Moreover, the noised image $x_t$ in Diff-AE incorporates noise added across the entire image. In contrast, in our approach, the noise is exclusively added to the masked region, thereby maintaining the unaltered section. Distinguished from $x_t$, the noised image here is denoted as $x^{\prime}_t$ here, as articulated in Equation \ref{modified_x_t}, where the symbol $\odot$ denotes element-wise product and $m$ designates a binary mask matrix: zero for the edited region and one for the unchanged part. This equation demonstrates that $m \odot x_0$ supplies the known pixel in the given image, and $(1-m) \odot x_t$ is a masked version of $x_t$ for each iteration $t$, providing the unknown pixel. The architecture is depicted in the first stage of Figure \ref{fig:arch_graph}.


\begin{equation}
\label{modified_x_t}
x^{\prime}_t=m \odot x_0+(1-m) \odot x_t
\end{equation}

Furthermore, it is crucial to observe that the facial image $x_f$ encompasses an additional eye area, extending beyond the masked region. The inclusion of this extra eye is anchored on the reasoning that it assists in localizing the position of the nose and mouth, thereby enhancing the stability of the nose and mouth across frames.

In the DDPM's training stage, we employ the simplified loss objective delineated in \cite{ho2020denoising} and incorporate a specific mask that ensures only the loss in the editable facial motion area is computed, as depicted in Equation \ref{loss}, where $\epsilon$ represents the actual noise. For inference, in light of the extensive iteration steps associated with diffusion, we opt for the Denoising Diffusion Implicit Model (DDIM) \cite{ddim}—an alternative non-Markovian noising process—as the solver to accelerate the sampling process.

\begin{equation}
\label{loss}
L_{\text {simple}}=E_{t, x_0, \epsilon}\left[\left\|(1-m) \odot (\epsilon-\epsilon_\theta\left(x^{\prime}_t, t, z_\text{sem}\right))\right\|^2\right]
\end{equation}

\subsection{Video Sequence Generation}

This phase aims to produce person-specific synthesized videos by processing $N$ reference facial images and singular driving audio. It enlists the help of a reference encoder and a motion generator based on the Conformer~\cite{gulati2020conformer}. Unlike methods~\cite{wav2lip, dinet} that restricted inputs to fixed frames, our methodology permits input lengths to vary. Incorporating the Conformer also enables us to capture global and local facial motion interactions, a considerable leap from previous methods \cite{wav2lip, pcavs, dinet} that only facilitated interactions of short durations.

Besides, in this stage, we rely on latent codes instead of predefined structural representations~\cite{fan2022faceformer, chen2023improving, iplap}. The latent codes fall into two categories: the semantic latent code $z$ and the audio latent code $a$. The semantic encoder, which remains frozen, is tasked with extracting the semantic latent code from facial images. Concurrently, the audio model, using a self-supervised approach, extracts the latent code from the driving audio. The reference encoder gleans person-specific facial texture information from $N$ visual latent codes $z_{\text{1:N}}$, and the motion generator performs a one-to-one mapping, transforming $T$ audio latent codes $a_{\text{1:T}}$ into the corresponding $T$ visual latent codes $z^{\prime}_{\text{1:T}}$. Ultimately, the visual latent codes are fed into the inpainting renderer to synthesize the images.

To efficiently incorporate personalized textures, we introduce a cross-attention mechanism~\cite{vaswani2017attention}, a pivot from the direct concatenation of the reference images~\cite{wav2lip, dinet, difftalk}. This mechanism employs the output of the reference decoder as the query, while the audio latent codes operate as the key and value in a multi-head attention operation, thus generating person-aware facial motion latent codes.

To enhance the robustness of the audio encoder, we retrieve the audio latent code through a weighted sum \cite{yang2021superb} of all layers of the self-supervised models. This approach deviates from the commonly used Mel-based feature representation, thus granting added language flexibility. Furthermore, mirroring the approach adopted in DAE-Talker~\cite{daetalker}, we use a shared $x_T$ for DDIM as the starting point for all images, thereby ensuring that the DDIM generates deterministic and consistent results.

\section{Experiments}
\label{sec:typestyle}

\subsection{Experimental Setups}

\textbf{Dataset.} 
We utilize the HDTF dataset \cite{hdtf} for our experiments. The HDTF dataset is comprised of high-resolution, real-world talking head videos collected from YouTube. The dataset has a total duration of around 16 hours, partitioning 245 clips for training and 68 clips for testing, aligning with the framework in \cite{dinet}. Notably, the actual tally of videos utilized in our experimental setup marginally trails the official release, owing to the unavailability of 6 online videos.
As part of the preprocessing endeavor, all videos in the HDTF dataset are reformatted to a fixed resolution of $256\times256$ pixels and standardized to a frame rate of 25 frames per second (FPS).

\textbf{Model Details.} Similar to Diff-AE~\cite{diffae}, the diffusion model draws on U-Net~\cite{ronneberger2022convolutional} and the dimension of the semantic latent code $\mathbf{z}_{\text{sem}}$ is 512. We employ several techniques to augment the data, including horizontal flipping, color jitter, Gaussian blur, shifting, scaling, and rotation. The time step $T$ for DDIM is set to 20.
The off-the-shelf audio encoder is a pre-trained Hubert-large model \cite{hsu2021hubert}. The reference frame number $N$ is set to 75~(3 seconds) for few-shot experiments. The reference encoder and motion generator use a 2-layer and 8-layer conformer, both employing two attention heads and relative positional encoding~\cite{dai2019transformer}. The model undergoes training for an initial four epochs in the first stage, followed by an additional 100 epochs in the second stage.

\textbf{Evaluation Metric.} As for \textbf{Visual Quality~(VQ)}, we employ Peak Signal-to-Noise Ratio~(PSNR), Structured similarity~(SSIM)~\cite{ssim} and Learned Perceptual Image Patch Similarity~(LPIPS)~\cite{lpips} as metrics to quantify the similarity between the generated and ground truth images. Since masked areas vary across methods, we resize all images to an identical resolution for a fair comparison and only evaluate the lower face area. As for \textbf{Synchronization~(SYNC)}, We utilize lip-sync-error distance (LSE-D), lip-sync-error confidence (LSE-C) \cite{wav2lip, syncnet}, and landmarks distance (LMD) \cite{xie2021towards}. The LSE metrics measure the degree of lip-sync alignment between the generated lips and the audio, while the LMD metric assesses the reconstructed shape of the lower face region relative to the ground truth.

\textbf{Baseline systems.} We compare our method with several state-of-the-art person-generic methods~\cite{wav2lip, pcavs, iplap, daetalker}. \textbf{Wav2Lip}~\cite{wav2lip} employs an auto-encoder trained through adversarial methods. \textbf{PC-AVS}~\cite{pcavs} introduces a pose-controllable audio-visual talking face generation method. \textbf{IP-LAP}~\cite{iplap} and \textbf{DAE-Talker}~\cite{daetalker} represent two-stage methodologies. IP-LAP employs landmark, whereas DAE-Talker utilizes latent code. For equal comparison, adjustments are made to accommodate the nature of each method. Since PC-AVS is influenced by pose, we utilize ground-truth pose information for its evaluation. Furthermore, due to the speaker-specific design of DAE-Talker, we engage in a retraining process on HDTF.

\subsection{Method Comparison}
We compare methods across three typical scenarios: {\em reconstruction}, {\em dubbing}, and {\em one-shot}. For the reconstruction and dubbing, we synthesized a talking head with audio and they differ in that the reconstruction employs ground-truth audio, while the dubbing utilizes audio from an alternate dialogue or other speakers. For one-shot, a single portrait is exploited to synthesize the outcome.


 \vspace{-10pt}

\begin{table}[!hpt]

  \caption[Dataset statistics]{Quantitative Results on HDTF Reconstruction}
  \label{tab:main_dataset_metrics}
  \centering
  \resizebox{1.0\linewidth}{!}{
  \begin{tabular}{@{}l cccccc@{}} \toprule
  \multirow{2}{*}{Method}  
    & \multicolumn{3}{c}{VQ} & \multicolumn{3}{c}{SYNC}  \\  \cmidrule(lr){2-4}  \cmidrule(lr){5-7} 
              & PSNR$\uparrow$ & SSIM$\uparrow$ & LPIPS$\downarrow$ & LSE-D$\downarrow$ & LSE-C$\uparrow$ & LMD$\downarrow$ \\  \cmidrule{1-7}

Wav2Lip~\cite{wav2lip} & 26.12& 	0.83 &	0.066 &	7.48 &	7.84 &	1.10   \\ 
PC-AVS~\cite{pcavs} & 22.15 &	0.68 &	0.058 &	\textbf{6.66} &	\textbf{8.69} &	2.03   \\ 
IP-LAP~\cite{iplap}  & 26.05 &	0.84 &	0.058 &	8.78 &	5.59 &	\textbf{0.79}  \\ 
DAE-Talker~\cite{daetalker} & 17.81 &	0.47 &	0.129 &	8.90 &	6.15 &	2.78 \\ 
\textit{GT} & \textit{N/A}&	\textit{1}	&\textit{0}&	\textit{6.73}&	\textit{9}	& \textit{0} \\  \cmidrule{1-7} 
\textbf{Ours}  & \textbf{28.18} &	\textbf{0.87} 	&\textbf{0.035} &	8.16 &	7.10 &	0.95    \\

    \bottomrule
  \end{tabular}
}
\end{table}

 \vspace{-10pt}
 
\textbf{Quantitative Comparison.} The results are recorded in Table~\ref{tab:main_dataset_metrics}. Our methodology exhibits commendable results in terms of the visual quality metric, corroborating the efficacy of the rendering module in the preliminary stage. However, when examined from a synchronization perspective, Wav2Lip and PC-AVS outpace our method across lip-sync-error (LSE) parameters, and IP-LAP surpasses our performance regarding the landmarks distance (LMD). This discrepancy stems from these methods' direct optimization focus on either LSE or LMD, a focus our method does not explicitly pursue. Nonetheless, our method delivers competitive results despite the absence of an explicit optimization for these distinct metrics.

\begin{figure}[!hpt]
    \centering
    \includegraphics[width=1.0\linewidth]{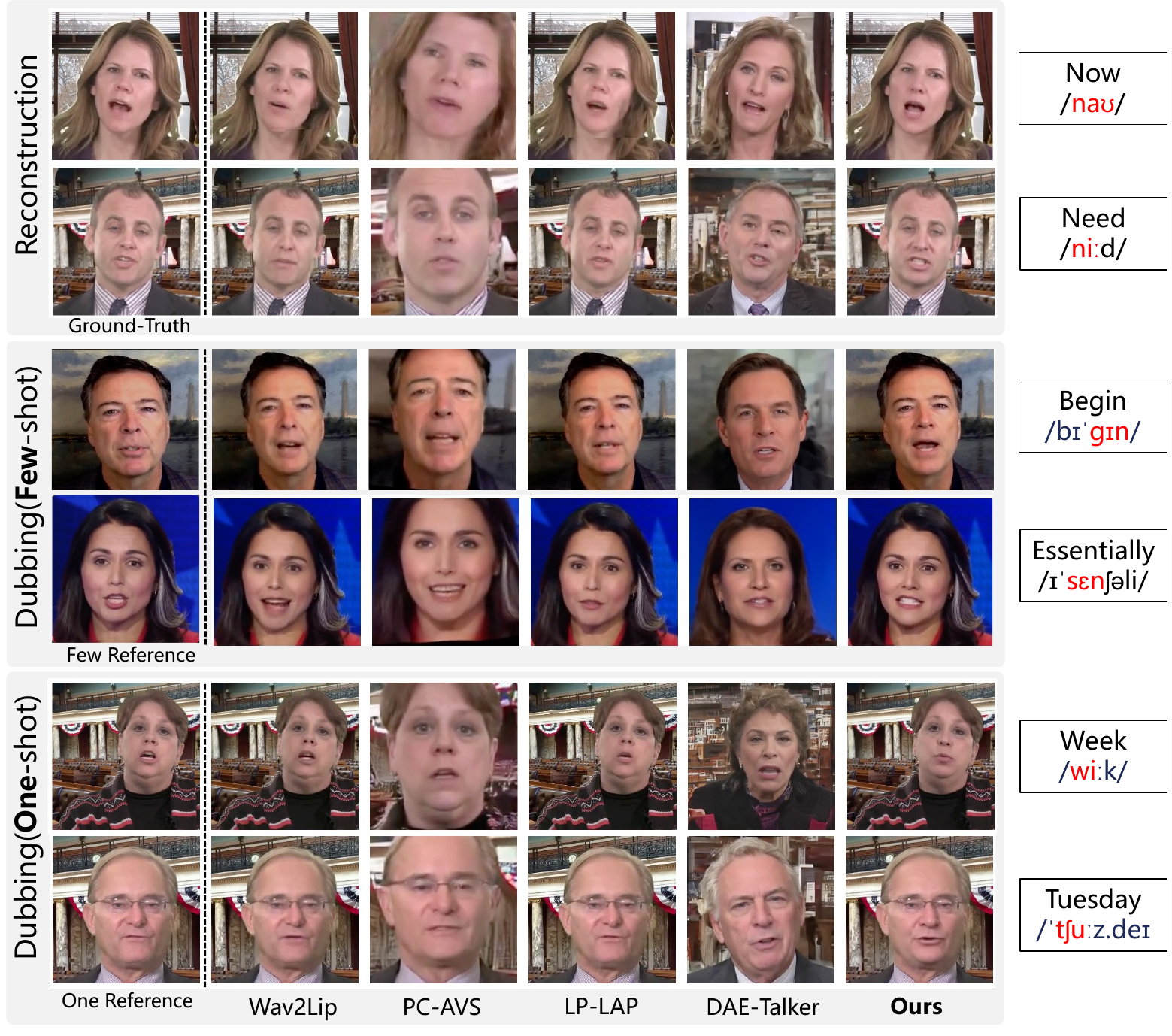}
        \caption{Qualitative results on Reconstruction \& Dubbing. The corresponding pronounced syllables are highlighted in red.}
    \label{fig:visualization}
\end{figure}

\textbf{Qualitative Comparison.} The results are presented in Figure \ref{fig:visualization}. It is noteworthy that the generated lower face and teeth regions manifest improved clarity, mirroring the positive visual quality metric unearthed in the quantitative comparison. Additionally, our generated outputs exhibit enhanced consistency with the longer pronunciation units such as syllables. This outcome reaffirms our method's capacity to produce more comprehensible results, synchronizing with the conclusion that previous strategies frequently culminate in lip-synced but unintelligible videos \cite{wang2023seeing}.

 \vspace{-10pt}

\begin{table}[!hpt]

  \caption[Dataset statistics]{Ablation Study on HDTF Reconstruction}
  \label{tab:ablation_study}
  \centering
  \resizebox{1.0\linewidth}{!}{
  \begin{tabular}{@{}l cccccc@{}} \toprule
  \multirow{2}{*}{Method}  
    & \multicolumn{3}{c}{VQ} & \multicolumn{3}{c}{SYNC}  \\  \cmidrule(lr){2-4}  \cmidrule(lr){5-7} 
              & PSNR$\uparrow$ & SSIM$\uparrow$ & LPIPS$\downarrow$ & LSE-D$\downarrow$ & LSE-C$\uparrow$ & LMD$\downarrow$ \\  \cmidrule{1-7} 

Ours w/o eye & 28.84 & 	0.88 &	0.033 &	8.85 &	5.76 &	1.53       \\ 
Ours w/o aug. & 28.45 &	0.88 &	0.034 &	9.60 &	4.75 &	0.94      \\
Ours w/o w.s. & 27.48  & 0.86	 &	0.040 &	8.40 & 6.50	 & 1.90  \\ 
Ours w $\frac{1}{10}$ data & 28.12 &	0.87 &	0.033 &	8.34 &	7.01 &	0.97 \\ \cmidrule{1-7} 
\textbf{Ours}  & 28.18 &	0.87	& 0.035 &	8.16 &	7.10 &	0.95    \\

    \bottomrule
  \end{tabular}
}
\end{table}

\textbf{Ablation Study.} We conducted several ablation experiments to evaluate the effectiveness of the proposed model. In the first ablation, we remove the eye area from the input for the semantic encoder. This exclusion led to a particular degradation in LMD, highlighting the importance of eyes in guiding the localization of the nose and mouth.
In the second ablation, we deactivated the augmentation in the first stage. The results revealed a reduction in LSE, suggesting that augmentation is crucial in learning high-level semantic information from facial images. In the third ablation, we adopted the Mel-based feature instead of weighted sum~(w.s.), and we observed a decline across all metrics.  In the fourth ablation, we curtailed the amount of paired audio-visual data utilized in the second stage to a tenth of the original data, equivalent to approximately 1.5 hours. Remarkably, we observe no significant degradation, implying that our method exhibits enhanced robustness with constrained training data.

 \vspace{-10pt}
 
\begin{table}[hbpt!]

  \caption[Dataset statistics]{User Study with Mean Option Score (MOS)}
  \label{tab:user_study_metrics}
  \centering
  \resizebox{0.9\linewidth}{!}{
  \begin{tabular}{@{}l |cccc@{}} \toprule
   Methods & MOS-MF &  MOS-RI & 	MOS-N  & MOS-CL\\  \hline

Wav2Lip~\cite{wav2lip} &  3.46 &	4.07 &	3.26  &  3.42  \\ 
PC-AVS~\cite{pcavs} &  3.74 &	3.46 &	3.08   & 3.50 \\ 
IP-LAP~\cite{iplap}  &  3.53 &	3.70 &	3.95 &	 2.42   \\ 
DAE-Talker~\cite{daetalker} &  3.19 &	3.33 &	2.61 &  2.23  \\  \hline
\textbf{Ours}  & \textbf{4.62} &	\textbf{4.59} &	\textbf{4.60}  & \textbf{4.17} \\ 

    \bottomrule
  \end{tabular}
}
\end{table}
 \vspace{-5pt}
 
\textbf{Subjective Evaluation.} We conducted a user study with 12 participants rating our method across Mouth Fidelity (MF), Reading Intelligibility (RI), Naturalness (N), and Cross-Lingual performance (CL). Five videos were selected randomly from HDTF dataset. For CL, the original videos were in English, and we translated them into Mandarin Chinese, Korean, and French. The results, outlined in Table \ref{tab:user_study_metrics}, demonstrate the language generality of our method.

 \vspace{-5pt}
 
\section{Conclusion}
\label{sec:conclusion}

 \vspace{-5pt}
This paper presents \textbf{\em{DiffDub}}, an innovative approach for person-generic dubbing without necessitating fine-tuning. We have devised a potent inpainting renderer with meticulously tailored strategies that can seamlessly integrate the editable region. We employ a conformer-based method with a cross-attention mechanism to learn person-specific textures and details, thus accommodating varied references and reducing dependence on the training corpus. Quantitative and qualitative experiments underline that our proposed method yields seamless and intelligible outcomes. Subjective evaluations further affirm that DiffDub outstrips the baseline by a margin.

 \vspace{-5pt}
 
\section{ACKNOWLEDGEMENTS}

 \vspace{-5pt}
This work was supported by Shanghai Municipal Science and Technology Major Project (No.~2021SHZDZX0102), the Key Research and Development Program of Jiangsu Province, China (No.~BE2022059), Scientific and Technological Innovation 2030 under Grant (No.~2021ZD0110900).


\vfill\pagebreak
\clearpage

\bibliographystyle{IEEEtran}
\bibliography{refs}

\end{document}